\pdfoutput=1

\documentclass[11pt]{article}

\usepackage[preprint]{acl}

\usepackage{times}
\usepackage{latexsym}

\usepackage[T1]{fontenc}

\usepackage[utf8]{inputenc}

\usepackage{microtype}

\usepackage{inconsolata}

\usepackage{graphicx}

\usepackage{times}
\usepackage{latexsym}
\usepackage{amsfonts}
\usepackage{amsmath}
\usepackage{subcaption}
\usepackage{graphicx}
\usepackage{pgfplots} 
\usepackage{scalefnt} 

\usepackage{graphicx}
\usepackage{xcolor}
\usepackage{soul}
\usepackage{CJKutf8}
\usepackage{tabularx}
\usepackage{booktabs}
\usepackage{makecell}
\usepackage[misc]{ifsym}
\usepackage{amssymb}
\usepackage{pifont}

\usepackage{arydshln} 
\usepackage{booktabs}

\usepackage{pgfplotstable} 

\usepackage[T1]{fontenc}

\usepackage[utf8]{inputenc}

\usepackage{microtype}

\usepackage{inconsolata}

\definecolor{color1}{RGB}{79,79,254}
\definecolor{color2}{RGB}{206,61,50}
\definecolor{color3}{RGB}{175,175,175}
\definecolor{color4}{RGB}{143,80,159}
\definecolor{blanc}{RGB}{255,255,255}

\definecolor{blueaccent}{RGB}{0,150,214}
\definecolor{greenaccent}{RGB}{0,139,43}
\definecolor{purpleaccent}{RGB}{130,41,128}
\definecolor{orangeaccent}{RGB}{240,83,50}

\newcommand{\revising}[1]{\textcolor{black}{#1}}

\newcommand{\white}[1]{\textcolor{blanc}{#1}}

\newcommand{\warningtext}[1]{\textcolor{red!80!black}{#1}}

\usepackage{multirow}
\usepackage{ulem}

\title{EpiCoDe: Boosting Model Performance Beyond Training with Extrapolation and Contrastive Decoding}

\author{ 
\textbf{Mingxu Tao\textsuperscript{1}},
 \textbf{Jie Hu\textsuperscript{2}},
 \textbf{Mingchuan Yang\textsuperscript{2}},
 \textbf{Yunhuai Liu\textsuperscript{3,4}},
\\
 \textbf{Dongyan Zhao\textsuperscript{1,5}},
 \textbf{Yansong Feng\textsuperscript{1\,\Letter}}
\\
 \textsuperscript{1}Wangxuan Institute of Computer Technology, Peking University
 \\
 \textsuperscript{2}Research Institute of China Telecom
 \\
 \textsuperscript{3}School of Computer Science, Peking University
 \\
 \textsuperscript{4}Beijing Institute of Big Data Research
 \\
 \textsuperscript{5}National Key Laboratory of General Artificial Intelligence, BIGAI
\\
\texttt{\{thomastao,fengyansong\}@pku.edu.cn} \\
}

\begin{document}
\maketitle
\begin{abstract}
The remarkable performance of Large language models (LLMs) relies heavily on the availability of abundant high-quality training data. 
However, 
the high cost of acquiring annotated data often prevents models from obtaining capabilities to tackle downstream tasks.
In this paper, we introduce a novel method, EpiCoDe that boosts model performance in data-scarcity scenarios without extra training. 
We first employ model extrapolation to enhance 
a finetuned model with its inferior 
version, and then adopt contrastive decoding to further reduce predicted errors, by comparing the logit scores given by the extrapolated and the vanilla finetuned model.  
Experiments across three tasks over four different LLMs show that EpiCoDe consistently outperforms existing methods with  significant and robust improvement. 
We also propose a new theoretical framework to reveal the mechanism behind contrastive decoding in data-scarcity scenarios, which further helps us better understand the effectiveness of EpiCoDe.   
\end{abstract}

\section{Introduction}


Large language models (LLMs) have demonstrated superior performance across a wide range of downstream tasks, making them an indispensable application in natural language processing~\cite{yang2024qwen2technicalreport,grattafiori2024llama3herdmodels,deepseekai2025deepseekr1}. 
Their impressive capabilities, however, are primarily dependent on the availability of large-scale, high-quality training data. 
The scarcity of such data, especially in specific domains, has become a major barrier to further advancing LLMs.
This challenge is particularly pronounced in fields such as law and health~\cite{huang2023lawyerllama,chen2024huatuogpto1}, where access to domain-specific data is often limited due to privacy concerns, copyright restrictions, and the high cost of data collection. 
Furthermore, the expense of high-quality annotations may also remain a bottleneck, restricting the consistent development of model abilities.

Therefore, there has been a growing need to develop methods that can make the most use of limited 
data 
without extra training to 
enhance model performance in such data-scarcity scenarios.
%
Existing efforts include two types of methods, model extrapolation~\cite{zheng2024weaktostrong} and contrastive decoding~\cite{li-etal-2023-contrastive}, which are developed from the perspectives of parameters and inference, respectively.
Model extrapolation collects two checkpoints from different training stages.
It obtains a stronger model by comparing the parameters of two checkpoints without extra training.
Beyond directly editing model parameters, there is another thread of research focusing on improving the probability to output correct next tokens during inference. 
\citeauthor{li-etal-2023-contrastive}~(\citeyear{li-etal-2023-contrastive}) hypothesize the LLMs trained on the same datasets can have similar patterns to make mistakes.
Thus, comparing the logit scores predicted by a small LLM and a large LLM in the same family and calculating their difference, contrastive decoding can reduce the shared errors of the two LLMs.

Previous work~\cite{karras2024guiding} also proposes to improve the quality of generated images by comparing the prediction of a model and its bad version trained with noisy data.
These two techniques are both good practice to enhance LLMs without extra training. 
However, they sometimes fail to bring improvement~\cite{obrien2023contrastive,zheng2024weaktostrong}. 
Furthermore, existing works have not explored whether they are effective in the context of data scarcity.

In this paper, we introduce EpiCoDe, a novel method that leverages model \underline{e}xtra\underline{p}olat\underline{i}on
and \underline{co}ntrastive \underline{de}coding 
to improve the performance of a finetuned model without further training in data-scarcity scenarios. 
We first obtain an extrapolated model of a finetuned LLM by employing its weakened version at early training stage as the anchor.
The resulting models as well as the vanilla finetuned model learn from the same dataset, exhibiting similar capabilities and error patterns.
We then compare their logit scores and employ contrastive decoding to further optimize the prediction of next tokens.
This frustratingly easy method can take the strengths of both methods to further improve model performance.

Previous efforts mainly provide empirical evidence to explain the success of contrastive decoding, without a proper analyzing tool to help us understand how contrastive decoding contributes to the improvement.
In this work, we further introduce a theoretical framework to explain the mechanisms behind contrastive decoding and investigate strategies for model selection in EpiCoDe. 
Our key insight is that 
model extrapolation helps maintain model locality across finetuned models and extrapolated models, which 
can ensure further improvement during the contrastive decoding stage. 
Specifically, we argue that EpiCoDe brings more stable improvement when selecting the weak models that are closer to the extrapolated model in the parameter space.

We evaluate our method 
across three distinct tasks (law, math, and logical reasoning) using four LLMs from different families and with various scales. 
Experimental results show that EpiCoDe consistently outperforms baseline models, which employ either model extrapolation or contrastive decoding, with 
significant performance gains across all tasks. 
Furthermore, our experiments also demonstrate the validity of our theoretical framework, highlighting the importance of model locality in the context of EpiCoDe.

Our main contributions are threefold:
\begin{itemize}
    \item We propose a frustratingly simple yet effective method, EpiCoDe, that benefits from both model extrapolation and contrastive decoding, 
    enhancing model capabilities without additional training, particularly in data-scarcity scenarios.
    \item We present a theoretical framework to explain the mechanism behind EpiCoDe and the model selection strategies in contrastive decoding, which have been only discussed empirically.  
    \item  Experiments on three diverse tasks across four different base models show that our method can bring consistent improvement, significantly outperforming baseline methods. 
\end{itemize}

\section{Related Works}

The scarcity of high-quality annotated training data often prevents us from effectively finetuning LLMs, necessitating the development of approaches that go beyond conventional training paradigms. Among others, 
model extrapolation and contrastive decoding are two representative threads of research falling to this type. 


\paragraph{Model Extrapolation} An intuitive idea to improve model performance is to extrapolate a model to a better position in the parameter space. 
However, we are unable to precisely locate the optimal point without training.
A simple approximate solution is to regard model extrapolation as the reverse process of linear merging~\cite{zheng2024weaktostrong}.
They first enhance a finetuned LLM with Reinforcement Learning with Human Feedback~(RLHF, \citealp{ouyang2022training}).
To obtain a better RLHF model, they hypothesize the RLHF model is constructed by merging the finetuned model and an unknown ultra-powerful model.
As shown in the left part of Figure~\ref{fig:method}, they argue that extrapolation can alleviate the limitation of insufficient training data, resulting a stronger RLHF model.

Here, we believe model extrapolation might be a possible choice to make the most of limited training data by exploring across different checkpoints.
In this paper, we extract an intermediate checkpoint during finetuning as the weak model. 
We then regard the model trained with more epochs as the strong model, and extrapolate it to obtain a stronger one. 

\paragraph{Contrastive Decoding} Different from model extrapolation which edits the parameters to get a new model, contrastive decoding~\cite{li-etal-2023-contrastive} combines two distinct LLMs' predictions during inference.
The fundamental hypothesis is that, on one hand, models trained on the same data may have similar patterns to make mistakes; while on the other hand, the capabilities of models differ due to model size.
\citeauthor{li-etal-2023-contrastive}~(\citeyear{li-etal-2023-contrastive}) proposes to collect the two probability distributions of next token predicted by a strong and a weak model in the same family, respectively. 
Their difference may represent the capability gap between two models without shared errors.
As shown in the right part of Figure~\ref{fig:method}, contrastive decoding improves the quality of model outputs by modifying the predicted logit scores with the computed differences.

In our work, we are interested in whether contrastive decoding can further enhance the extrapolated models. 
Different from previous works~\cite{li-etal-2023-contrastive,obrien2023contrastive} that use LLMs in the same family but with different sizes, we here adopt the finetuned model and the extrapolated model for exploration. 
The effectiveness of contrastive decoding in data-scarcity scenarios, e.g., enhancing a finetuned model with its early version, is still an unexplored research question.

\section{Methodology}



In this paper, we propose EpiCoDe, which leverages model extrapolation and contrastive decoding in data-scarcity scenarios, to boost model performance without additional training. 



\subsection{Preliminary}




In data scarcity scenarios, the training data available for finetuning is insufficient. We can only obtain under-trained LLMs with limited capabilities. 
And, model checkpoints collected from the earlier stage of such training conditions can perform unsurprisingly worse. 

\paragraph{{Model Extrapolation.}} 
Following previous work~\cite{zheng2024weaktostrong}, we regard a strong model $\theta^{strong}$ as the linearly merged model~\cite{pmlr-v162-wortsman22a,choshen2022fusing} of its weakened version $\theta^{weak}$ and an ultra-strong model.
Thus, we can obtain the unknown model by linear extrapolation:
$$\theta^{ep}=\theta^{strong}+\mu(\theta^{strong}-\theta^{weak}),$$
where $\mu>0$ is a hyper-parameter. 

\label{sec:cd_define}


\paragraph{{Contrastive Decoding}} For the input sequence $x_{<i}:\,\left<x_1,\,\cdots,\,x_{i-1}\right>$, our aim is to predict the next token $x_i$ within the vocabulary $\mathcal{V}$. 
Given the distribution $p_s(x_i|x_{<i})$ predicted by a strong model and $p_w(x_i|x_{<i})$ predicted by a weak model, previous work~\cite{li-etal-2023-contrastive} defines the \textit{contrastive difference} as:
$$c(x_i|x_{<i})=\log\left(p_s(x_i|x_{<i})\right)-\log\left(p_w(x_i|x_{<i})\right),$$
which excludes the shared errors of the two models. 

Let $L_s(x_{<i})$ and $L_w(x_{<i})\in\mathbb{R}^{\mathcal{V}}$ denote the unnormalized distributions~(i.e., logit scores) predicted by the strong and the weak model. 
We can reformulate the \textit{contrastive difference} as:
$$c(x_{<i})=L_s(x_{<i})-L_w(x_{<i}).$$

Following previous work~\cite{obrien2023contrastive}, we further define the logit score for next token as:
$$L_{CD}(x_{<i}) = L_{s}(x_{<i})+\lambda c(x_{<i}),$$ 
where $\lambda>0$ is a hyper-parameter controlling the strength of contrastive decoding.




\subsection{Our Method: EpiCoDe}

\begin{figure}[t]
\centering
\includegraphics[width=1\columnwidth]{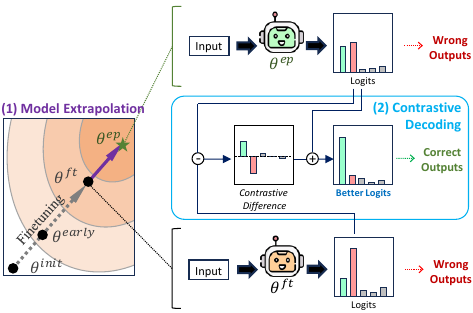}
\caption{The process of EpiCoDe, which first employs model extrapolation to obtain a stronger LLM $\theta^{ep}$, and then uses contrastive decoding to get better logit scores.}
\label{fig:method}
\end{figure}

%
Formally, we denote the initial LLM as $\theta^{init}$ and finetune it on the training dataset.
We can collect the model from two training stages:
(1) the model $\theta^{early}$ at an early stage, which may be finetuned with part of the training data or trained with fewer epochs; (2) the model $\theta^{ft}$ finetuned on all data with more epochs.

However, in data scarcity cases, the model $\theta^{ft}$ may be still undertrained, due to limited training data available.
To address this challenge, we develop 
a novel method EpiCoDe, which integrates model extrapolation and contrastive decoding, without the need of any additional training.
Figure~\ref{fig:method} illustrates the process of EpiCoDe.

We first employ model extrapolation, where $\theta^{early}$ serves as the weak model and $\theta^{ft}$ as the strong model, to obtain the extrapolated model $\theta^{ep}=\theta^{ft}+\mu(\theta^{ft}-\theta^{early})$.
By selecting an appropriate $\mu$, we can derive an extrapolated model $\theta^{ep}$ with better capabilities compared to $\theta^{ft}$. 

During inference, we use contrastive decoding for better performance. 
As shown in the left part of Figure~\ref{fig:method}, we have several models in hand, whose performance should have following ranking: $\theta^{init}<\theta^{early}<\theta^{ft}<\theta^{ep}$.
We intuitively choose $\theta^{ep}$ as the strong model in contrastive decoding, while one of the other three models can serve as the weak model.
An important research question is which model should be employed to achieve the best performance. 
We argue that the similar error patterns between the weak and the strong model play an important role in the effectiveness of contrastive decoding.
Thus, we employ $\theta^{ft}$, which locates the nearest to $\theta^{ep}$ in the parameter space, as the weak model.
The \textit{contrastive difference} can be refined as $c(x_{<i})=L_{ep}(x_{<i})-L_{ft}(x_{<i})$, with $L_{ep}(x_{<i})$ and $L_{ft}(x_{<i})$ denoting the logit scores yield by $\theta^{ep}$ and $\theta^{ft}$, respectively.
Similarly, we represent the logit scores in contrastive decoding as $L_{CD}=L_{ep}(x_{<i})+\lambda c(x_{<i})$.

Our pilot experiments show that using $\theta^{ft}$ as the weak model in contrastive decoding can result in good performance~(See details in Section~\ref{sec:locality_discuss}). 
We keep this setting for main experiments. 

\section{Why can EpiCoDe Work?}
\label{sec:local_theory}

Previous analysis~\cite{chang-etal-2024-explaining} explains that contrastive decoding may play a role similar to reduce the loss.
However, a theoretical framework for understanding the underlying mechanisms still remains absent.

Different from previous work~\cite{li-etal-2023-contrastive,obrien2023contrastive}, in the contrastive decoding of EpiCoDe, we use models from different stages based on the same LLM, rather than models of different sizes.
Our models vary in capabilities due to the quantity of finetuning data or whether they are obtained through model extrapolation. 

In the context of EpiCoDe, we are uncertain whether using the extrapolated model $\theta^{ep}$ and its inferior version for contrastive decoding can be effective.
With $\theta^{ep}$ serving as the strong model, we also aim to study which other model should be employed as the weak model.
Thus, we develop a theoretical toolkit to explain the role of contrastive decoding in EpiCoDe and how to select the weak model yielding optimal performance.

\subsection{Error Analysis of Contrastive Decoding}

In this paper, we propose the theory from the perspective of logit scores.
%
Let $\theta^{*}$ denote the hypothetical optimal model which may learn from a very large amount data, while $\theta$ denotes a model trained on insufficient data~(e.g., our models $\theta^{init}$, $\theta^{early}$, $\theta^{ft}$, and $\theta^{ep}$). 
Given the input sequence $x_{<i}$ of any test example, we formulate the error in logit score predicted by $\theta$ as:
$$\delta (x_{<i},\,\theta)=L(x_{<i}|\theta)-L(x_{<i}|\theta^{*}).$$

In contrastive decoding, we employ a strong $\theta^{s}$ and a weak model $\theta^{w}$ to yield logit scores, respectively.
We suppose that $\delta(*,\,\theta^{s})$ follows a normal distribution $\mathcal{N}(0,\,\epsilon^2)$.
Similarly, the error of weak model's logit scores has $\delta(*,\,\theta^{w})\sim\mathcal{N}\left(0,(k\epsilon)^2\right)$, with $k>1$.

We can then re-formulate logit score of contrastive decoding in Section~\ref{sec:cd_define} as:
\begin{multline*}
L_{CD}(x_{<i})=L(x_{<i}|\theta^{*})+\\
(1+\lambda)\delta (x_{<i},\,\theta^{s})-\lambda\delta (x_{<i},\,\theta^{w}).
\end{multline*}


The variance of the new logit score under contrastive decoding depends on the correlation between $\delta(*,\,\theta^{s})$ and $\delta(*,\,\theta^{w})$. 
We primarily study two typical scenarios: (1) The strong and the weak have learned similar capabilities; (2) The two model are not trained on the same data.

\subsection{Locality in EpiCoDe}
The variance of logit error can approach its lower bound $\left(1-\lambda(k-1)\right)\epsilon$, when $\delta(*,\,\theta^{s})$ and $\delta(*,\,\theta^{w})$ exhibit perfect positive correlation\footnote{Denote the probability density of $\delta(*,\,\theta^{s})$ and $\delta(*,\,\theta^{w})$ as $f_{s}(u)$ and $f_w(u)$, respectively, where $u \in \mathbb{R}^{\mathcal{V}}$. Here, we should have $f_w(u)=kf_s(u)$ for any $u$. }
This is an ideal scenario when the strong and the weak model have consistent patterns to make mistakes. 

We argue that $\theta^{ep}$ and $\theta^{ft}$ naturally have \textbf{locality}, ensuring their similar error patterns.
Given the evaluating loss on downstream tasks $\mathcal{L}$, the improvement of model extrapolation is $\Delta\mathcal{L}=\mathcal{L}\left(\theta^{ep}\right)-\mathcal{L}\left(\theta^{ft}\right)$.
To ensure model extrapolation can bring positive effects, we have to limit $\mu\ll 1$ and estimate the improvement by $\Delta\mathcal{L}=\mu(\theta^{ep}-\theta^{ft})\cdot\nabla\mathcal{L}(\theta^{ft}).$
Since LLMs suffer from insufficient training, this inner product can usually keep less than zero. 


Note that $\theta^{ep}$ lies within the neighborhood of $\theta^{ft}$.
The \textbf{locality} in model extrapolation ensures them with similar capabilities and patterns to make mistakes.
Therefore, we believe that employing $\theta^{ft}$ as the weak model can align closely with this idealized scenario.
Contrastive decoding can help to decrease the variance of errors in the predicted logit scores (by up to $\lambda(k-1)\epsilon$).


\subsection{Negative Effect of Inconsistent Errors}
We further introduce another idealized scenario where the strong and the weak model learn from different data.
The two models make mistakes inconsistently, and we can suppose that $\delta(*,\,\theta^{s})$ and $\delta(*,\,\theta^{w})$ are independent of each other.
In this case, the error of the logit scores in contrastive decoding follows a normal distribution $\mathcal{N}\left(0,\,(1+\lambda)^2\epsilon^2+\lambda^2k^2\epsilon^2\right)$.
This means the logit scores given by contrastive decoding even deviate more from the hypothetical optimal model $\theta^{*}$.
The initial model $\theta^{init}$ may hardly learn the same capabilities of $\theta^{ep}$.
Thus, we argue that using $\theta^{init}$ as the weak model may result in worse performance.

\subsection{Weak Model Selection}
Building on the two scenarios discussed above, we now analyze the effect of using $\theta^{early}$ as the weak model, which hinges on two key factors: (1) the performance gap between $\theta^{early}$ and $\theta^{ep}$, which affects $k$; and (2) the correlation between the errors in the logit scores predicted by $\theta^{early}$ and $\theta^{ep}$, determining whether contrastive decoding can successfully bring improvement.

Previous works~\cite{obrien2023contrastive,chang-etal-2024-explaining} propose to use smaller LLMs as the weak model.
Our theory supports such findings, since a larger $k$ helps to degrade the lower bound of error variance $(1-\lambda(k-1))\epsilon$.
Note that the LLMs with different sizes are trained on sufficient data, resulting in their similar capabilities and error patterns.
We argue that, however, in data-scarcity scenarios, several updating steps can significantly separate $\theta^{ft}$ from $\theta^{early}$ in the parameter space, which violates
the \textbf{locality} between $\theta^{early}$ and $\theta^{ep}$.

We argue that using $\theta^{early}$ as the weak model will make the logit error higher than lower bound, and may even result in negative effects like the second idealized scenario.
Thus, in EpiCoDe, we consistently choose $\theta^{ft}$ as the weak model in our main experiments. 
We provide more experimental results in Section~\ref{sec:locality_discuss} to examine our theory.

\section{Experimental Settings}

\paragraph{Datasets} We evaluate our proposed method in three different tasks that usually lack finetuning data: law, mathematics, and logical reasoning.
In the legal domain, we employ JEC-QA~\cite{zhong2019jecqa} as the training set and DISC-Law-Eval~\cite{yue2023disclawllm} as the test set. 
Following previous work~\cite{huang2023lawyerllama}, we convert these multiple-choice legal questions into true-or-false questions.
For mathematics and logical reasoning, we use GSM8K\_zh~\cite{yu2024metamath} and LogiQA~\cite{logiqa} for finetuning and evaluation, respectively.

\begin{table}[t]
\small
\centering
\begin{tabular}{lcccc}
\toprule
\multirow{2}{*}{\textbf{Task}} & \multicolumn{3}{c}{\textbf{Number of Examples}} & \multirow{2}{*}{\textbf{Avg. Len.}} \\
\cmidrule{2-4}
 & Train & Dev & Test & \\
\midrule
Law & 17,664 & 5,126 & 5,126 & 265.3 \\
Math & 7,471 & 660 & 659 & 117.2\\
Logic & 12,672 & 1,000 & 1,000 & 480.4 \\

\bottomrule
\end{tabular}
\caption{Statistics of the finetuning and evaluation datasets for each task. \textbf{Avg. Len.} shows average output length of training examples, counted by character.}
\label{tab:dataset_stat}
\end{table}

For the legal question answering~(QA) and logical reasoning tasks, which require multiple reasoning steps to handle, we collect chain-of-thought~(CoT, \citealp{wei2022chain}) data to finetune LLMs.
For the mathematical task, we utilize the intermediate solution steps from GSM8K\_zh for training, despite their brevity.
Table~\ref{tab:dataset_stat} summarizes overall statistics of the datasets.
We search the optimal hyper-parameters on hold-out validation sets before testing. See more details in Appendix~\ref{sec:appendixa}.

\paragraph{Metrics} Given that our models have undergone finetuning, we evaluate their performance in a zero-shot setting. 
We adopt accuracy for legal QA and logical reasoning tasks, while exact match for the mathematical task.

\paragraph{Models} We explore the effectiveness of EpiCoDe on the LLMs from three families with different scales, including Deepseek-7B-Chat~\cite{deepseekllm}, Qwen2-1.5B-Instruct, Qwen2-7B-Instruct~\cite{yang2024qwen2technicalreport}, and Llama-3.2-3B-Instruct~\cite{grattafiori2024llama3herdmodels}.
We continually finetune the instruct models with our data for two epochs.
We regard the checkpoint after one epoch as $\theta^{early}$, while the checkpoint after two epochs as $\theta^{ft}$.
Here we use AdamW~\cite{loshchilov2018decoupled} as the optimizer, with $\beta_1$ and $\beta_2$ are set to 0.9 and 0.95 respectively. The maximum learning rate is set to 3e-5, and the batch size is set to 128.

\paragraph{Decoding Constraints} 
Contrastive decoding has potential flaws in computing the logit scores of incorrect tokens. 
Specifically, if the weak model assigns an extremely low logit score to a wrong token $e\in\mathcal{V}$, the \textit{contrastive difference} of this token $c(x_i=e|x_{<i})$ becomes excessively large.
The model enhanced with contrastive decoding can be more prone to generating this incorrect token.

To address this issue, following previous work~\cite{li-etal-2023-contrastive}, we introduce a threshold $\alpha$, which restricts the model must select the next token from those with high logit scores predicted by the strong model.


For all experiments, we empirically set $\alpha$ to 0.1.




\section{Main Results}

We adopt the models finetuned for two epochs as baseline. 
As comparative experiments, we also study the improvements when using only model extrapolation~\cite{zheng2024weaktostrong} or contrastive decoding~\cite{obrien2023contrastive}. 
For model extrapolation, we directly evaluate the performance of  $\theta^{ep}$. 
And for contrastive decoding, we employ $\theta^{early}$ and $\theta^{ft}$ as the weak and the strong model, respectively.
We repeat to train the LLMs on each dataset for 10 times with different random seeds \revising{and report the average accuracy}.

\begin{table}[t]
\small
\tabcolsep=0.55em
\centering
\begin{tabular}{lcccc}
\toprule
\multirow{2}{*}{\textbf{Method}} & \multicolumn{3}{c}{\textbf{Accuracy$_{\text{(}\Delta\text{Acc)}}$}} & \multirow{2}{*}{\textbf{Avg.}}\\
\cmidrule{2-4}
& Law & Math & Logic & \\
\midrule
\multicolumn{5}{l}{\textit{Deepseek-7B-Chat}} \\
Finetune & 
    64.78$_{\white{\text{(+0.00)}}}$ & 27.28$_{\white{\text{(+0.00)}}}$ & 57.22$_{\white{\text{(+0.00)}}}$ & 49.76\\
\hdashline
ME & 
    65.42$_{\text{(+0.64)}}$ & 27.12$_{\text{(\ -0.17)}}$ & 58.89$_{\text{(+1.67)}}$ & 50.48\\
CD &
    65.29$_{\text{(+0.51)}}$ & 26.88$_{\text{(\ -0.41)}}$ & 58.46$_{\text{(+1.24)}}$ & 50.21\\
EpiCoDe &
    \textbf{65.51}$_{\text{(+0.73)}}$ & \textbf{27.81}$_{\text{(+0.53)}}$ & \textbf{59.05}$_{\text{(+1.83)}}$ & \textbf{50.79}\\
\midrule
\multicolumn{5}{l}{\textit{Qwen2-1.5B-Instruct}} \\
Finetune & 
    64.34$_{\white{\text{(+0.00)}}}$ & 39.48$_{\white{\text{(+0.00)}}}$ & 52.68$_{\white{\text{(+0.00)}}}$ & 52.17\\
\hdashline
ME & 
    64.94$_{\text{(+0.60)}}$ & 40.80$_{\text{(+1.32)}}$ & 53.63$_{\text{(+0.95)}}$ & 53.12\\
CD &
    65.22$_{\text{(+0.88)}}$ & 39.23$_{\text{(\ -0.26)}}$ & 53.42$_{\text{(+0.74)}}$ & 52.62\\
EpiCoDe &
    \textbf{65.38}$_{\text{(+1.04)}}$ & \textbf{41.20}$_{\text{(+1.71)}}$ & \textbf{53.87}$_{\text{(+1.19)}}$ & \textbf{53.48}\\
\midrule
\multicolumn{5}{l}{\textit{Qwen2-7B-Instruct}} \\
Finetune & 
    69.03$_{\white{\text{(+0.00)}}}$ & 57.12$_{\white{\text{(+0.00)}}}$ & 66.67$_{\white{\text{(+0.00)}}}$ & 64.27\\
\hdashline
ME & 
    69.46$_{\text{(+0.43)}}$ & 57.79$_{\text{(+0.67)}}$ & 67.60$_{\text{(+0.93)}}$ & 64.95\\
CD &
    69.91$_{\text{(+0.88)}}$ & 58.62$_{\text{(+1.50)}}$ & 67.43$_{\text{(+0.76)}}$ & 65.32\\
EpiCoDe &
    \textbf{70.25}$_{\text{(+1.22)}}$ & \textbf{58.71}$_{\text{(+1.59)}}$ & \textbf{68.07}$_{\text{(+1.40)}}$ & \textbf{65.68}\\
\midrule
\multicolumn{5}{l}{\textit{Llama-3.2-3B-Instruct}} \\
Finetune & 
    62.13$_{\white{\text{(+0.00)}}}$ & 48.45$_{\white{\text{(+0.00)}}}$ & 53.45$_{\white{\text{(+0.00)}}}$ & 54.68\\
\hdashline
ME & 
    62.73$_{\text{(+0.60)}}$ & 49.74$_{\text{(+1.29)}}$ & 55.11$_{\text{(+1.66)}}$ & 55.77\\
CD &
    63.38$_{\text{(+1.25)}}$ & 53.13$_{\text{(+4.68)}}$ & 56.62$_{\text{(+3.17)}}$ & 57.71\\
EpiCoDe &
    \textbf{63.79}$_{\text{(+1.66)}}$ & \textbf{54.31}$_{\text{(+5.86)}}$ & \textbf{57.48}$_{\text{(+4.03)}}$ & \textbf{58.53}\\
\bottomrule
\end{tabular}
\caption{Accuracy of distinct LLMs at each stage on the three tasks. We abbreviate contrastive decoding as CD and model extrapolation as EP.}
\label{tab:main_results}
\end{table}


As shown in Table~\ref{tab:main_results}, model extrapolation and contrastive decoding are both effective methods to enhance LLMs without training. 
In general, using only model extrapolation results in an average improvement of 0.68\%\textasciitilde1.09\% across the three tasks, while contrastive decoding yields improvement of 0.45\%\textasciitilde3.03\%.
However, \textbf{our proposed EpiCoDe outperforms these two baselines}, enhancing the finetuned models with the most improvement of 1.03\%\textasciitilde3.85\%. 

We find that the effects of model extrapolation and contrastive decoding vary on different LLMs. 
For instance, contrastive decoding can bring significant improvement~(1.25\%\textasciitilde4.68\%) to Llama-3.2-3B-Instruct, which is trained on fewer Chinese data, while model extrapolation enhance the accuracy by merely 0.60\%\textasciitilde1.66\%.
However, for the other LLMs which have learned from sufficient Chinese texts, model extrapolation is usually more effective than contrastive decoding. 
Differently, \textbf{EpiCoDe can always leverage the strengths of both methods}, consistently yielding superior improvements across all LLMs.

\begin{table}[t]
\small
\centering
\begin{tabular}{llccc}
\toprule
\multirow{2}{*}{\textbf{$\mathcal{H}_0$}} & \multirow{2}{*}{\textbf{Model}} & \multicolumn{3}{c}{\textbf{Significance Level $\alpha$}} \\
\cmidrule{3-5}
& & Law & Math & Logic \\
\midrule
\multirow{4}{*}{EpiCoDe$>$ME} & \textit{DS-7B} & \warningtext{0.0899} & 0.0198 & \warningtext{0.0821} \\
& \textit{QW-1.5B} & 0.0026 & \warningtext{0.0938} & \warningtext{0.0711} \\
& \textit{QW-7B} & 0.0000 & 0.0067 & \warningtext{0.0783} \\
& \textit{LM-3B} & 0.0000 & 0.0000 & 0.0001 \\
\midrule
\multirow{4}{*}{EpiCoDe$>$CD} & \textit{DS-7B} & 0.0190 & 0.0341 & 0.0267 \\
& \textit{QW-1.5B} & 0.0342 & 0.0001 & 0.0116 \\
& \textit{QW-7B} & 0.0279 & \warningtext{0.3344} & 0.0130 \\
& \textit{LM-3B} & 0.0113 & 0.0007 & 0.0006 \\
\bottomrule
\end{tabular}
\caption{Significance level of mistakenly rejecting the null hypothesis $\mathcal{H}_0$ in one-tailed tests. The \warningtext{dark red} scores mean that the confidence to believe $\mathcal{H}_0$ being true is lower than \textbf{95\%}. DS-7B, QW-1.5B, QW-7B, and LM-3B represent to employ Deepseek-7B-Chat, Qwen2-1.5B-Instruct, Qwen2-7B-
Instruct, and Llama-3.2-3B-Instruct as the initial LLM, respectively.}
\label{tab:significance_test}
\end{table}

\paragraph{\revising{Significance of EpiCoDe's Improvement}} \revising{We further study whether EpiCoDe can bring significantly more improvement than the baseline methods. 
For each task and each random seed, we evaluate the performance of three settings: (1)using only model extrapolation, (2) using only contrastive decoding, and (3) our proposed EpiCoDe method. Then, we compare the performance of the EpiCoDe method with the two baseline methods under the same random seed, conducting a paired t-test. The null hypothesis is that \textit{EpiCoDe performs better than each of the two baseline methods}.}

\revising{Table~\ref{tab:significance_test} illustrates the magnitudes of the significance level $\alpha$, which means the probability of mistakenly rejecting the null hypothesis.
Except for Qwen2-7B-Instruct on Math, we can conclude with more than 90\% confidence level to believe that EpiCoDe outperforms the two baseline methods, using only model extrapolation or only contrastive decoding.
Furthermore, across the 4 LLMs on 3 tasks, we have over 95\% confidence to believe that EpiCoDe surpasses model extrapolation in 7~(/12) experiments and outperforms contrastive decoding in 11~(/12) experiments.
Thus, we believe \textbf{EpiCoDe is significantly better than the two baselines}, which achieves optimal performance.}

\begin{table}[t]
\small
\centering
\begin{tabular}{lcccc}
\toprule
\multirow{2}{*}{\textbf{Method}} & \multicolumn{3}{c}{\textbf{Num. of Successful Runs}} & \multirow{2}{*}{\makecell[c]{\textbf{Total}\\(*/30)}}\\
\cmidrule{2-4}
& Law & Math & Logic \\
\midrule
\multicolumn{4}{l}{\textit{Deepseek-7B-Chat}} \\
ME & 8 & 6 & \textbf{10} & 24\\
CD & 8 & 4 & 8  & 20\\
EpiCoDe & \textbf{10} & \textbf{8} & \textbf{10} & \textbf{28} \\
\midrule
\multicolumn{4}{l}{\textit{Qwen2-1.5B-Instruct}} \\
ME & 8 & 9 & 7 & 24\\
CD & 9 & 4 & 7 & 20\\
EpiCoDe & \textbf{10} & \textbf{10} & \textbf{9} & \textbf{29}\\
\midrule
\multicolumn{4}{l}{\textit{Qwen2-7B-Instruct}} \\
ME & 8 & 8 & 8 & 24\\
CD & \textbf{10} & 8 & 7 & 25 \\
EpiCoDe & \textbf{10} & \textbf{10} & \textbf{10} & \textbf{30} \\
\midrule
\multicolumn{4}{l}{\textit{Llama-3.2-3B-Instruct}} \\
ME & \textbf{10} & \textbf{10} & \textbf{10} & \textbf{30} \\
CD &\textbf{10} & \textbf{10} & \textbf{10} & \textbf{30} \\
EpiCoDe & \textbf{10} & \textbf{10} & \textbf{10} & \textbf{30} \\
\bottomrule
\end{tabular}
\caption{The number of times that each method brings improvement to the finetuned models across 10 runs.}
\label{tab:num_of_success}
\end{table}

\paragraph{\revising{Robustness of EpiCoDe}} Table~\ref{tab:main_results} also shows that EpiCoDe can consistently enhance all the four LLMs across three tasks.
But in the mathematical task, \textbf{using only contrastive decoding or only model extrapolation sometimes fails} to bring improvement~(e.g., for Deepseek-7B-Chat and Qwen2-1.5B-Instruct). 
Thus, we further examine the number of times that each method successfully enhances the finetuned model across 10 repeated experiments.
As shown in Table~\ref{tab:num_of_success}, \textbf{our EpiCoDe can provide very robust improvement for the finetuned LLMs}, with merely three failures in a total of 120 experiments.
As a comparison, model extrapolation yields improvement in 80\% of the experiments on Deepseek-7B-Chat and the Qwen2 family. 
Even worse, contrastive decoding exhibits a higher frequency to fail in Deepseek-7B-Chat and Qwen2-1.5B-Instruct, especially on the mathematical task, where it has positive effects to each model merely 4 times.
But for Qwen2-7B-Instruct, the two baselines have comparable rates to achieve better accuracy than finetuned models~(24 vs. 25 out of 30 runs).
Note that as a very strong LLM, Qwen2-7B-Instruct may have learned from various datasets, including those similar to our tasks. 
We guess that its parameters change less during the training process, which ensures \textbf{locality} between $\theta^{early}$ and $\theta^{ft}$, an essential premise for contrastive learning to work. See more discussion about locality in Section~\ref{sec:locality_discuss}.

For 10 runs across all three tasks, EpiCoDe and the two baselines can consistently improve Llama-3.2-3B-Instruct, which suffers extremely from the lack of Chinese training data.
Since model extrapolation and contrastive decoding are both effective, we believe that \textbf{the combination of two techniques plays a vital role to boost model performance in scenarios with restricted training resources.}

\section{Discussions}
\subsection{Where does the improvement come from?}

Table~\ref{tab:main_results} shows that the effectiveness of each method varies across tasks.
For instance, in logical reasoning, all three methods can improve the performance of each finetuned LLM, while in legal QA, the improvement becomes relatively less.
However, in the mathematical task, only EpiCoDE yields improvement for all LLMs, while using only model extrapolation or only contrastive decoding even degrades performance for models such as Qwen2-1.5B-Instruct and Llama-3.2-3B-Instruct.

We guess that \textbf{the difficulty of a task may influence the effectiveness of these methods}. 
Specifically, GSM8K\_zh is a simple task, primarily consisting of grade school math problems. 
In contrast, legal QA and logical reasoning, as more difficult tasks, require LLMs to possess enough domain-specific knowledge and commonsense, to distinguish similar concepts, and to apply multi-step reasoning.
The average length of training examples in Table~\ref{tab:dataset_stat} can represent the difficulties of each task~(longer outputs for the more complex task).

To dive deeper into where EpiCoDe brings the improvement, we take legal QA as an example. 
For each run, we collect the token sequences outputted by the finetuned model~($\theta^{ft}$), and count their lengths.
We partition the test set of legal QA into three equal parts according to the length order.
We then evaluate the improvement of the three methods on each subset.

\begin{figure*}[t] 
\centering 
\begin{subfigure}{0.5\columnwidth}
\resizebox{1\columnwidth}{!}{  
    \pgfplotstableread[col sep=comma]{
	Year, Law, Math, Logic
	1, 0.41, -1.65, 0.18
	2, 0.56, -0.18, 1.17
	3, 0.73, 0.53, 1.83
	}\mytablea

\begin{tikzpicture}
  \begin{axis}[
    width=1.8\columnwidth, height=1.8\columnwidth,
    ybar,
    bar width=7.5pt,
    ymin=-1.8,ymax=2,
    enlarge x limits={abs=25pt},
    legend style={draw=none,at={(0.5,-0.2)},
    anchor=north,legend columns=-1},
    symbolic x coords={1,2,3},
    xtick={1,2,3},
    xticklabels={{\LARGE$\theta^{init}$},{\LARGE$\theta^{early}$},{\LARGE$\theta^{ft}$}},
    ytick={-1,0,1,2},
    yticklabels={-1,\white{-}0,1,2},
    cycle list={blueaccent,greenaccent,purpleaccent}
  ]
    \pgfplotsinvokeforeach{Law,Math,Logic}{
      \addplot+[draw=none,fill,] table[x=Year,y=#1]{\mytablea};
      \addlegendentry{#1};
    }
    
\end{axis}
\end{tikzpicture}
}
\caption{Deepseek-7B-Chat}
\label{fig:ds}
\end{subfigure}
\begin{subfigure}{0.5\columnwidth}
\resizebox{1\columnwidth}{!}{  
    \pgfplotstableread[col sep=comma]{
	Year, Law, Math, Logic
	1, 0.38, -0.64, -0.23
	2, 0.62, -0.01, 1.02
	3, 1.04, 1.71, 1.17
	}\mytableb

\begin{tikzpicture}
  \begin{axis}[
    width=1.8\columnwidth, height=1.8\columnwidth,
    ybar,
    bar width=7.5pt,
    ymin=-0.75,ymax=1.8,
    enlarge x limits={abs=25pt},
    legend style={draw=none,at={(0.5,-0.2)},
    anchor=north,legend columns=-1},
    symbolic x coords={1,2,3},
    xtick={1,2,3},
    xticklabels={{\LARGE$\theta^{init}$},{\LARGE$\theta^{early}$},{\LARGE$\theta^{ft}$}},
    ytick={0,1},
    yticklabels={\white{-}0,1},
    cycle list={blueaccent,greenaccent,purpleaccent}
  ]
    \pgfplotsinvokeforeach{Law,Math,Logic}{
      \addplot+[draw=none,fill,] table[x=Year,y=#1]{\mytableb};
      \addlegendentry{#1}
    }
\end{axis}
\end{tikzpicture}
}
\caption{Qwen2-1.5B-Instruct}
\label{fig:qw15}
\end{subfigure}
\begin{subfigure}{0.5\columnwidth}
\resizebox{1\columnwidth}{!}{  
    \pgfplotstableread[col sep=comma]{
	Year, Law, Math, Logic
	1, -0.33, -1.16, 0.11
	2, 0.60, 1.15, 0.08
	3, 1.22, 1.59, 1.40
	}\mytableb

\begin{tikzpicture}
  \begin{axis}[
    width=1.8\columnwidth, height=1.8\columnwidth,
    ybar,
    bar width=7.5pt,
    ymin=-1.27,ymax=1.7,
    enlarge x limits={abs=25pt},
    legend style={draw=none,at={(0.5,-0.2)},
    anchor=north,legend columns=-1},
    symbolic x coords={1,2,3},
    xtick={1,2,3},
    xticklabels={{\LARGE$\theta^{init}$},{\LARGE$\theta^{early}$},{\LARGE$\theta^{ft}$}},
    ytick={-1,0,1},
    yticklabels={-1,\white{-}0,1},
    cycle list={blueaccent,greenaccent,purpleaccent}
  ]
    \pgfplotsinvokeforeach{Law,Math,Logic}{
      \addplot+[draw=none,fill,] table[x=Year,y=#1]{\mytableb};
      \addlegendentry{#1}
    }
\end{axis}
\end{tikzpicture}
}
\caption{Qwen2-7B-Instruct}
\label{fig:qw7}
\end{subfigure}
\begin{subfigure}{0.5\columnwidth}
\resizebox{1\columnwidth}{!}{  
    \pgfplotstableread[col sep=comma]{
	Year, Law, Math, Logic
	1, -0.14, 0.66, -0.68
	2, 0.75, 2.29, 3.11
	3, 1.66, 5.86, 4.03
	}\mytableb

\begin{tikzpicture}
  \begin{axis}[
    width=1.8\columnwidth, height=1.8\columnwidth,
    ybar,
    bar width=7.5pt,
    ymin=-0.9,ymax=6.05,
    enlarge x limits={abs=25pt},
    legend style={draw=none,at={(0.5,-0.2)},
    anchor=north,legend columns=-1},
    symbolic x coords={1,2,3},
    xtick={1,2,3},
    xticklabels={{\LARGE$\theta^{init}$},{\LARGE$\theta^{early}$},{\LARGE$\theta^{ft}$}},
    ytick={0,2,4,6},
    yticklabels={\white{-}0,2,4,6},
    cycle list={blueaccent,greenaccent,purpleaccent}
  ]
    \pgfplotsinvokeforeach{Law,Math,Logic}{
      \addplot+[draw=none,fill,] table[x=Year,y=#1]{\mytableb};
      \addlegendentry{#1}
    }
\end{axis}
\end{tikzpicture}
}
\caption{Llama-3.2-3B-Instruct}
\label{fig:lm3}
\end{subfigure}
\caption{
The improvement of using $\theta^{init}$, $\theta^{early}$, or $\theta^{ft}$ as the weak model, compared to the performance of vanilla finetuned model $\theta^{init}$.} 
\label{fig:local_check_results}  
\end{figure*}
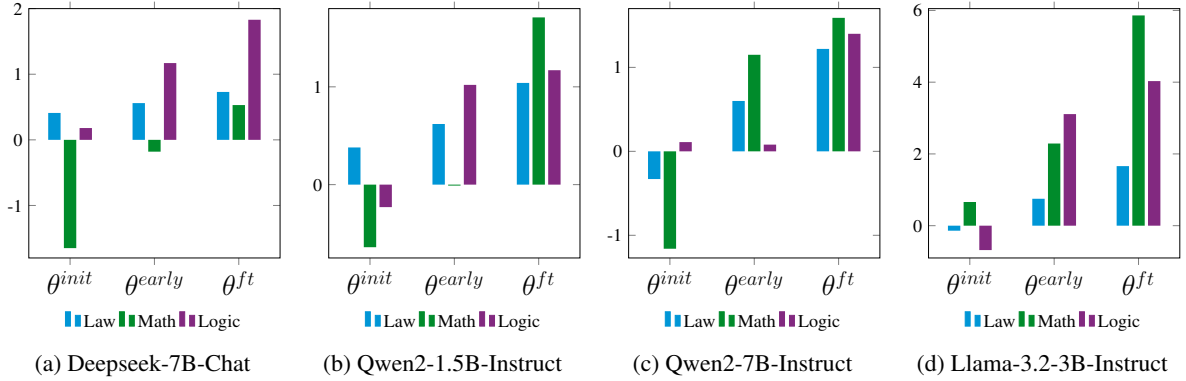

\begin{table}[t]
\small
\tabcolsep=0.4em
\centering
\sethlcolor{yellow}
\begin{tabular}{lcccccc}
\toprule
\multirow{2}{*}{\textbf{Method}} & \multicolumn{2}{c}{\textbf{Easy}} & \multicolumn{2}{c}{\textbf{Medium}} & \multicolumn{2}{c}{\textbf{Hard}} \\
\cmidrule(lr){2-3} \cmidrule(lr){4-5} \cmidrule(lr){6-7}
& Acc. & $\Delta$Acc. & Acc. & $\Delta$Acc. & Acc. & $\Delta$Acc. \\
\midrule
\multicolumn{5}{l}{\textit{Deepseek-7B-Chat}} \\
Finetune & 68.81 & -- & 65.08 & -- & 60.65 & -- \\
ME & 68.86 & +0.05 & 65.26 & \textit{+0.18} & 62.31 & {\textbf{+1.67}} \\
EpiCoDe & 68.93 & +0.13 & 65.28 & \textit{+0.19} & 62.49 & {\textbf{+1.84}}\\
\hdashline
CD only & 68.62 & \ -0.18 & 65.16 & \textit{+0.08} & 62.25 & {\textbf{+1.60}} \\
\midrule
\multicolumn{5}{l}{\textit{Qwen2-1.5B-Instruct}} \\
Finetune & 69.05 & -- & 64.37 & -- & 59.68 & --\\
ME & 69.31 & +0.26 & 64.89 & \textit{+0.52} & 60.70 & {\textbf{+1.02}} \\
EpiCoDe & 69.38 & +0.33 & 65.88 & {\textbf{+1.52}} & 60.96 & \textit{+1.28} \\
\hdashline
CD only & 69.25 & +0.20 & 65.40 & \textit{+1.03} & 61.07 & {\textbf{+1.39}} \\
\midrule
\multicolumn{5}{l}{\textit{Qwen2-7B-Instruct}} \\
Finetune & 75.77 & -- & 69.19 & -- & 62.36 & -- \\
ME & 75.68 & \ -0.10 & 69.52 & \textit{+0.33} & 63.40 & {\textbf{+1.04}} \\
EpiCoDe & 75.70 & \ -0.07 & 70.48 & \textit{+1.28} & 64.80 & {\textbf{+2.44}} \\
\hdashline
CD only & 75.65 & \ -0.13 & 69.95 & \textit{+0.76} & 64.34 & {\textbf{+1.98}} \\
\midrule
\multicolumn{5}{l}{\textit{Llama-3.2-3B-Instruct}} \\
Finetune & 64.74 & -- & 62.36 & -- & 59.37 & -- \\ 
ME & 65.15 & +0.41 & 63.07 & {\textbf{+0.71}} & 60.04 & \textit{+0.67} \\
EpiCoDe & 65.27 & +0.52 & 63.84 & \textit{+1.48} & 62.33 & {\textbf{+2.96}} \\
\hdashline
CD only & 65.23 & +0.48 & 63.58 & \textit{+1.21} & 61.39 & {\textbf{+2.03}} \\
\bottomrule
\end{tabular}
\caption{Each method's accuracy on the three subsets of legal QA. For each experiment, \revising{we compare the improvement brought by corresponding method on the three subsets. T}he highest improvement among three subsets is made \textbf{bold}, while the second highest is made \textit{italic}.}
\label{tab:improve_on_each_shard}
\end{table}

Table~\ref{tab:improve_on_each_shard} shows the performance of vanilla finetuned LLMs and the models enhanced by each method. 
By comparing the results of finetuned models and extrapolated models, we find that \textbf{model extrapolation can significantly benefits the hard subsets}~(needing the longest CoT), resulting in improvement of +1.02\%\textasciitilde+1.67\% on Qwen family and Deepseek-7B-Chat. 
But for the easy part, model extrapolation improves the performance much less, sometimes even degrading the accuracy.
However, for Llama-3.2-3B-Instruct which has limited Chinese capabilities, model extrapolation brings comparable improvement to three subsets.

We then examine the performance of EpiCoDe.
EpiCoDe surpasses model extrapolation by +0.17\%\textasciitilde+1.40\% on the hard subset, while merely +0.01\textasciitilde+1.00\% on the medium subset.
Thus, our EpiCoDe, which is further enhanced by contrastive decoding, also gains improvement primarily from the hard part.

We also study the effect of using only contrastive decoding, where $\theta^{ft}$ and $\theta^{early}$ serve as the strong and the weak model, respectively. 
It also results in the most improvement~(+1.39\%\textasciitilde+2.03\%) on the hard subsets, but relatively less improvement~(+0.08\%\textasciitilde+1.21\%) on the medium ones and even nearly no improvement on the easy ones.

We believe that model extrapolation and contrastive decoding play similar roles in improving LLMs' capabilities, \textbf{especially when generating longer CoT to tackle challenging problems}.
We also argue that \textbf{EpiCoDe can integrate the strengths of both techniques to achieve optimal performance.}


\subsection{Examination of Locality}
\label{sec:locality_discuss}

After model extrapolation, we can obtain four models, $\theta^{init}$, $\theta^{early}$, $\theta^{ft}$, and $\theta^{ep}$.
We intuitively employ $\theta^{ep}$ as the strong model in contrastive decoding, due to its strongest capabilities.
We aim to study the effect of selecting each of the other models as the weak model.

Figure~\ref{fig:local_check_results} illustrates the improvement of employing each model as the weak model during contrastive decoding.
We find that when using the initial model $\theta^{init}$ as the weak model, the performance of EpiCoDe often lags behind the vanilla finetuned model $\theta^{ft}$, especially in mathematical tasks. 
This demonstrates that \textbf{the weak model must learn from the same datasets to the extrapolated model; otherwise, contrastive decoding will have a negative effect}, even neutralizing the improvements brought by model extrapolation.

Previous works~\cite{obrien2023contrastive,chang-etal-2024-explaining} argue that a larger capability gap between the weak and the strong model can bring more improvement, which indicates $\theta^{early}$ may be a better choice than $\theta^{ft}$.
However, Figure~\ref{fig:local_check_results} shows that, if we employ $\theta^{early}$ as the weak model in contrastive decoding, EpiCoDe often achieves comparable accuracy to the extrapolated model $\theta^{ep}$, for instance, Deepseek-7B-Chat and Qwen2-1.5B on legal QA and logical reasoning. 
In the mathematical task, using $\theta^{early}$ also makes the improvement of EpiCoDe down to around zero.
As shown in Section~\ref{sec:local_theory}, we empirically demonstrate that 
\textbf{in data-scarcity scenarios, several update steps may break the locality of models, resulting in inconsistent error patterns}.

Comparing the results in Figure~\ref{fig:local_check_results}, we find that employing $\theta^{ft}$ as the weak model can consistently achieve the optimal performance for each LLM across all tasks.
We are confident that when selecting models from different training stages for contrastive decoding, \textbf{similar error patterns between the strong and the weak model is more crucial than a large capability gap}. 
Therefore, different from previous works~\cite{obrien2023contrastive,chang-etal-2024-explaining}, the best strategy here is to choose models from adjacent stages.

\section{Conclusion}

In this paper, we propose an easy yet effective method, EpiCoDe, to enhance finetuned LLMs in data scarcity scenarios.  
Benefiting from both model extrapolation and contrastive decoding, EpiCoDe can operate without additional training. 
Extensive experiments across different tasks show that EpiCoDe can bring consistent improvement, significantly outperforming baseline models. 
%
%
We also propose a theoretical framework from the perspective of logit errors to analyze the mechanisms of how EpiCoDe can work better.
We find that model extrapolation naturally ensures the locality between the finetuned LLM and the extrapolated LLM. 
During inference, their locality further fulfills the prerequisite that enables contrastive decoding to be effective.
%

\section*{Limitations}
\paragraph{Investigated Tasks} Due to the limitation of copyrights, 
there are still no publicly accessible legal QA or logical testing datasets in English. 
Thus, we employ the legal QA datasets adapted from the judicial examination of China and the logical reasoning dataset from National Public Service Examination in China.

Such tasks are challenging enough, since they require LLMs to possess domain-specific knowledge and commonsense and to obtain final answers through complex multi-step reasoning.
Existing LLMs still can not perform well on such tasks~\cite{yue2023disclawllm,logiqa20}, which aligns the need of developing training-free methods to enhance model capabilities.

\paragraph{Investigated Models} Due to the restrictions of computational resources, we mainly study the LLMs with scales from 1.5B to 7B. 
The results of LLMs from difference families across all three tasks show that our proposed EpiCoDe can consistently outperform the baselines.
We believe that EpiCoDe can be effective for larger LLMs, which suffer more from data scarcity.

\section*{Acknowledgments}
\revising{
This work is supported in part by NSFC~(62161160339) and Beijing Science and Technology Program~(Z231100007423011).
We thank Di He, Xiao Liu, Huishuai Zhang, and the anonymous reviewers for their helpful discussions and suggestions. 
For any correspondence, please contact Yansong Feng.
}

\bibliography{custom,anthology}

\begin{thebibliography}{21}
\providecommand{\natexlab}[1]{#1}

\bibitem[{Bi et~al.(2024)Bi, Chen, Chen, Chen, Dai, Deng, Ding, Dong, Du, Fu, Gao, Gao, Gao, Ge, Guan, Guo, Guo, Hao, Hao, He, Hu, Huang, Li, Li, Li, Li, Li, Liang, Lin, Liu, Liu, Liu, Liu, Liu, Liu, Lu, Lu, Luo, Ma, Nie, Pei, Piao, Qiu, Qu, Ren, Ren, Ruan, Sha, Shao, Song, Su, Sun, Sun, Tang, Wang, Wang, Wang, Wang, Wang, Wu, Wu, Xie, Xie, Xie, Xiong, Xu, Xu, Xu, Yang, You, Yu, Yu, Zhang, Zhang, Zhang, Zhang, Zhang, Zhang, Zhang, Zhang, Zhao, Zhao, Zhou, Zhou, Zhu, and Zou}]{deepseekllm}
Xiao Bi, Deli Chen, Guanting Chen, Shanhuang Chen, Damai Dai, Chengqi Deng, Honghui Ding, Kai Dong, Qiushi Du, Zhe Fu, Huazuo Gao, Kaige Gao, Wenjun Gao, Ruiqi Ge, Kang Guan, Daya Guo, Jianzhong Guo, Guangbo Hao, Zhewen Hao, Ying He, Wenjie Hu, Panpan Huang, Erhang Li, Guowei Li, Jiashi Li, Yao Li, Y.~K. Li, Wenfeng Liang, Fangyun Lin, A.~X. Liu, Bo~Liu, Wen Liu, Xiaodong Liu, Xin Liu, Yiyuan Liu, Haoyu Lu, Shanghao Lu, Fuli Luo, Shirong Ma, Xiaotao Nie, Tian Pei, Yishi Piao, Junjie Qiu, Hui Qu, Tongzheng Ren, Zehui Ren, Chong Ruan, Zhangli Sha, Zhihong Shao, Junxiao Song, Xuecheng Su, Jingxiang Sun, Yaofeng Sun, Minghui Tang, Bingxuan Wang, Peiyi Wang, Shiyu Wang, Yaohui Wang, Yongji Wang, Tong Wu, Y.~Wu, Xin Xie, Zhenda Xie, Ziwei Xie, Yiliang Xiong, Hanwei Xu, R.~X. Xu, Yanhong Xu, Dejian Yang, Yuxiang You, Shuiping Yu, Xingkai Yu, B.~Zhang, Haowei Zhang, Lecong Zhang, Liyue Zhang, Mingchuan Zhang, Minghua Zhang, Wentao Zhang, Yichao Zhang, Chenggang Zhao, Yao Zhao, Shangyan Zhou, Shunfeng Zhou, Qihao Zhu,
  and Yuheng Zou. 2024.
\newblock \href {https://arxiv.org/abs/2401.02954} {Deepseek llm: Scaling open-source language models with longtermism}.
\newblock \emph{Preprint}, arXiv:2401.02954.

\bibitem[{Chang et~al.(2024)Chang, Peng, Bansal, Ramakrishna, and Chung}]{chang-etal-2024-explaining}
Haw-Shiuan Chang, Nanyun Peng, Mohit Bansal, Anil Ramakrishna, and Tagyoung Chung. 2024.
\newblock \href {https://doi.org/10.18653/v1/2024.emnlp-main.484} {Explaining and improving contrastive decoding by extrapolating the probabilities of a huge and hypothetical {LM}}.
\newblock In \emph{Proceedings of the 2024 Conference on Empirical Methods in Natural Language Processing}, pages 8503--8526, Miami, Florida, USA. Association for Computational Linguistics.

\bibitem[{Chen et~al.(2024)Chen, Cai, Ji, Wang, Liu, Wang, Hou, and Wang}]{chen2024huatuogpto1}
Junying Chen, Zhenyang Cai, Ke~Ji, Xidong Wang, Wanlong Liu, Rongsheng Wang, Jianye Hou, and Benyou Wang. 2024.
\newblock \href {https://arxiv.org/abs/2412.18925} {Huatuogpt-o1, towards medical complex reasoning with llms}.
\newblock \emph{Preprint}, arXiv:2412.18925.

\bibitem[{Choshen et~al.(2022)Choshen, Venezian, Slonim, and Katz}]{choshen2022fusing}
Leshem Choshen, Elad Venezian, Noam Slonim, and Yoav Katz. 2022.
\newblock \href {https://arxiv.org/abs/2204.03044} {Fusing finetuned models for better pretraining}.
\newblock \emph{Preprint}, arXiv:2204.03044.

\bibitem[{DeepSeek-AI et~al.(2025)DeepSeek-AI, Guo, Yang, Zhang, Song, Zhang, Xu, Zhu, Ma, Wang, Bi, Zhang, Yu, Wu, Wu, Gou, Shao, Li, Gao, Liu, Xue, Wang, Wu, Feng, Lu, Zhao, Deng, Zhang, Ruan, Dai, Chen, Ji, Li, Lin, Dai, Luo, Hao, Chen, Li, Zhang, Bao, Xu, Wang, Ding, Xin, Gao, Qu, Li, Guo, Li, Wang, Chen, Yuan, Qiu, Li, Cai, Ni, Liang, Chen, Dong, Hu, Gao, Guan, Huang, Yu, Wang, Zhang, Zhao, Wang, Zhang, Xu, Xia, Zhang, Zhang, Tang, Li, Wang, Li, Tian, Huang, Zhang, Wang, Chen, Du, Ge, Zhang, Pan, Wang, Chen, Jin, Chen, Lu, Zhou, Chen, Ye, Wang, Yu, Zhou, Pan, Li, Zhou, Wu, Ye, Yun, Pei, Sun, Wang, Zeng, Zhao, Liu, Liang, Gao, Yu, Zhang, Xiao, An, Liu, Wang, Chen, Nie, Cheng, Liu, Xie, Liu, Yang, Li, Su, Lin, Li, Jin, Shen, Chen, Sun, Wang, Song, Zhou, Wang, Shan, Li, Wang, Wei, Zhang, Xu, Li, Zhao, Sun, Wang, Yu, Zhang, Shi, Xiong, He, Piao, Wang, Tan, Ma, Liu, Guo, Ou, Wang, Gong, Zou, He, Xiong, Luo, You, Liu, Zhou, Zhu, Xu, Huang, Li, Zheng, Zhu, Ma, Tang, Zha, Yan, Ren, Ren, Sha, Fu, Xu, Xie, Zhang,
  Hao, Ma, Yan, Wu, Gu, Zhu, Liu, Li, Xie, Song, Pan, Huang, Xu, Zhang, and Zhang}]{deepseekai2025deepseekr1}
DeepSeek-AI, Daya Guo, Dejian Yang, Haowei Zhang, Junxiao Song, Ruoyu Zhang, Runxin Xu, Qihao Zhu, Shirong Ma, Peiyi Wang, Xiao Bi, Xiaokang Zhang, Xingkai Yu, Yu~Wu, Z.~F. Wu, Zhibin Gou, Zhihong Shao, Zhuoshu Li, Ziyi Gao, Aixin Liu, Bing Xue, Bingxuan Wang, Bochao Wu, Bei Feng, Chengda Lu, Chenggang Zhao, Chengqi Deng, Chenyu Zhang, Chong Ruan, Damai Dai, Deli Chen, Dongjie Ji, Erhang Li, Fangyun Lin, Fucong Dai, Fuli Luo, Guangbo Hao, Guanting Chen, Guowei Li, H.~Zhang, Han Bao, Hanwei Xu, Haocheng Wang, Honghui Ding, Huajian Xin, Huazuo Gao, Hui Qu, Hui Li, Jianzhong Guo, Jiashi Li, Jiawei Wang, Jingchang Chen, Jingyang Yuan, Junjie Qiu, Junlong Li, J.~L. Cai, Jiaqi Ni, Jian Liang, Jin Chen, Kai Dong, Kai Hu, Kaige Gao, Kang Guan, Kexin Huang, Kuai Yu, Lean Wang, Lecong Zhang, Liang Zhao, Litong Wang, Liyue Zhang, Lei Xu, Leyi Xia, Mingchuan Zhang, Minghua Zhang, Minghui Tang, Meng Li, Miaojun Wang, Mingming Li, Ning Tian, Panpan Huang, Peng Zhang, Qiancheng Wang, Qinyu Chen, Qiushi Du, Ruiqi Ge, Ruisong
  Zhang, Ruizhe Pan, Runji Wang, R.~J. Chen, R.~L. Jin, Ruyi Chen, Shanghao Lu, Shangyan Zhou, Shanhuang Chen, Shengfeng Ye, Shiyu Wang, Shuiping Yu, Shunfeng Zhou, Shuting Pan, S.~S. Li, Shuang Zhou, Shaoqing Wu, Shengfeng Ye, Tao Yun, Tian Pei, Tianyu Sun, T.~Wang, Wangding Zeng, Wanjia Zhao, Wen Liu, Wenfeng Liang, Wenjun Gao, Wenqin Yu, Wentao Zhang, W.~L. Xiao, Wei An, Xiaodong Liu, Xiaohan Wang, Xiaokang Chen, Xiaotao Nie, Xin Cheng, Xin Liu, Xin Xie, Xingchao Liu, Xinyu Yang, Xinyuan Li, Xuecheng Su, Xuheng Lin, X.~Q. Li, Xiangyue Jin, Xiaojin Shen, Xiaosha Chen, Xiaowen Sun, Xiaoxiang Wang, Xinnan Song, Xinyi Zhou, Xianzu Wang, Xinxia Shan, Y.~K. Li, Y.~Q. Wang, Y.~X. Wei, Yang Zhang, Yanhong Xu, Yao Li, Yao Zhao, Yaofeng Sun, Yaohui Wang, Yi~Yu, Yichao Zhang, Yifan Shi, Yiliang Xiong, Ying He, Yishi Piao, Yisong Wang, Yixuan Tan, Yiyang Ma, Yiyuan Liu, Yongqiang Guo, Yuan Ou, Yuduan Wang, Yue Gong, Yuheng Zou, Yujia He, Yunfan Xiong, Yuxiang Luo, Yuxiang You, Yuxuan Liu, Yuyang Zhou, Y.~X. Zhu,
  Yanhong Xu, Yanping Huang, Yaohui Li, Yi~Zheng, Yuchen Zhu, Yunxian Ma, Ying Tang, Yukun Zha, Yuting Yan, Z.~Z. Ren, Zehui Ren, Zhangli Sha, Zhe Fu, Zhean Xu, Zhenda Xie, Zhengyan Zhang, Zhewen Hao, Zhicheng Ma, Zhigang Yan, Zhiyu Wu, Zihui Gu, Zijia Zhu, Zijun Liu, Zilin Li, Ziwei Xie, Ziyang Song, Zizheng Pan, Zhen Huang, Zhipeng Xu, Zhongyu Zhang, and Zhen Zhang. 2025.
\newblock \href {https://arxiv.org/abs/2501.12948} {Deepseek-r1: Incentivizing reasoning capability in llms via reinforcement learning}.
\newblock \emph{Preprint}, arXiv:2501.12948.

\bibitem[{Huang et~al.(2023)Huang, Tao, Zhang, An, Jiang, Chen, Wu, and Feng}]{huang2023lawyerllama}
Quzhe Huang, Mingxu Tao, Chen Zhang, Zhenwei An, Cong Jiang, Zhibin Chen, Zirui Wu, and Yansong Feng. 2023.
\newblock \href {https://arxiv.org/abs/2305.15062} {Lawyer llama technical report}.
\newblock \emph{Preprint}, arXiv:2305.15062.

\bibitem[{Karras et~al.(2024)Karras, Aittala, Kynk{\"a}{\"a}nniemi, Lehtinen, Aila, and Laine}]{karras2024guiding}
Tero Karras, Miika Aittala, Tuomas Kynk{\"a}{\"a}nniemi, Jaakko Lehtinen, Timo Aila, and Samuli Laine. 2024.
\newblock \href {https://openreview.net/forum?id=bg6fVPVs3s} {Guiding a diffusion model with a bad version of itself}.
\newblock In \emph{The Thirty-eighth Annual Conference on Neural Information Processing Systems}.

\bibitem[{Li et~al.(2023)Li, Holtzman, Fried, Liang, Eisner, Hashimoto, Zettlemoyer, and Lewis}]{li-etal-2023-contrastive}
Xiang~Lisa Li, Ari Holtzman, Daniel Fried, Percy Liang, Jason Eisner, Tatsunori Hashimoto, Luke Zettlemoyer, and Mike Lewis. 2023.
\newblock \href {https://doi.org/10.18653/v1/2023.acl-long.687} {Contrastive decoding: Open-ended text generation as optimization}.
\newblock In \emph{Proceedings of the 61st Annual Meeting of the Association for Computational Linguistics (Volume 1: Long Papers)}, pages 12286--12312, Toronto, Canada. Association for Computational Linguistics.

\bibitem[{Liu et~al.(2023)Liu, Liu, Cui, Teng, Duan, Zhou, and Zhang}]{logiqa20}
Hanmeng Liu, Jian Liu, Leyang Cui, Zhiyang Teng, Nan Duan, Ming Zhou, and Yue Zhang. 2023.
\newblock \href {https://doi.org/10.1109/TASLP.2023.3293046} {Logiqa 2.0—an improved dataset for logical reasoning in natural language understanding}.
\newblock \emph{IEEE/ACM Transactions on Audio, Speech, and Language Processing}, 31:2947--2962.

\bibitem[{Liu et~al.(2020)Liu, Cui, Liu, Huang, Wang, and Zhang}]{logiqa}
Jian Liu, Leyang Cui, Hanmeng Liu, Dandan Huang, Yile Wang, and Yue Zhang. 2020.
\newblock \href {https://doi.org/10.24963/ijcai.2020/501} {Logiqa: A challenge dataset for machine reading comprehension with logical reasoning}.
\newblock In \emph{Proceedings of the Twenty-Ninth International Joint Conference on Artificial Intelligence, {IJCAI-20}}, pages 3622--3628. International Joint Conferences on Artificial Intelligence Organization.
\newblock Main track.

\bibitem[{Loshchilov and Hutter(2019)}]{loshchilov2018decoupled}
Ilya Loshchilov and Frank Hutter. 2019.
\newblock \href {https://openreview.net/forum?id=Bkg6RiCqY7} {Decoupled weight decay regularization}.
\newblock In \emph{International Conference on Learning Representations}.

\bibitem[{O'Brien and Lewis(2023)}]{obrien2023contrastive}
Sean O'Brien and Mike Lewis. 2023.
\newblock \href {https://arxiv.org/abs/2309.09117} {Contrastive decoding improves reasoning in large language models}.
\newblock \emph{Preprint}, arXiv:2309.09117.

\bibitem[{Ouyang et~al.(2022)Ouyang, Wu, Jiang, Almeida, Wainwright, Mishkin, Zhang, Agarwal, Slama, Gray, Schulman, Hilton, Kelton, Miller, Simens, Askell, Welinder, Christiano, Leike, and Lowe}]{ouyang2022training}
Long Ouyang, Jeffrey Wu, Xu~Jiang, Diogo Almeida, Carroll Wainwright, Pamela Mishkin, Chong Zhang, Sandhini Agarwal, Katarina Slama, Alex Gray, John Schulman, Jacob Hilton, Fraser Kelton, Luke Miller, Maddie Simens, Amanda Askell, Peter Welinder, Paul Christiano, Jan Leike, and Ryan Lowe. 2022.
\newblock \href {https://openreview.net/forum?id=TG8KACxEON} {Training language models to follow instructions with human feedback}.
\newblock In \emph{Advances in Neural Information Processing Systems}.

\bibitem[{Team(2024)}]{grattafiori2024llama3herdmodels}
Llama Team. 2024.
\newblock \href {https://arxiv.org/abs/2407.21783} {The llama 3 herd of models}.
\newblock \emph{Preprint}, arXiv:2407.21783.

\bibitem[{Wei et~al.(2022)Wei, Wang, Schuurmans, Bosma, brian ichter, Xia, Chi, Le, and Zhou}]{wei2022chain}
Jason Wei, Xuezhi Wang, Dale Schuurmans, Maarten Bosma, brian ichter, Fei Xia, Ed~H. Chi, Quoc~V Le, and Denny Zhou. 2022.
\newblock \href {https://openreview.net/forum?id=_VjQlMeSB_J} {Chain of thought prompting elicits reasoning in large language models}.
\newblock In \emph{Advances in Neural Information Processing Systems}.

\bibitem[{Wortsman et~al.(2022)Wortsman, Ilharco, Gadre, Roelofs, Gontijo-Lopes, Morcos, Namkoong, Farhadi, Carmon, Kornblith, and Schmidt}]{pmlr-v162-wortsman22a}
Mitchell Wortsman, Gabriel Ilharco, Samir~Ya Gadre, Rebecca Roelofs, Raphael Gontijo-Lopes, Ari~S Morcos, Hongseok Namkoong, Ali Farhadi, Yair Carmon, Simon Kornblith, and Ludwig Schmidt. 2022.
\newblock \href {https://proceedings.mlr.press/v162/wortsman22a.html} {Model soups: averaging weights of multiple fine-tuned models improves accuracy without increasing inference time}.
\newblock In \emph{Proceedings of the 39th International Conference on Machine Learning}, volume 162 of \emph{Proceedings of Machine Learning Research}, pages 23965--23998. PMLR.

\bibitem[{Yang et~al.(2024)Yang, Yang, Hui, Zheng, Yu, Zhou, Li, Li, Liu, Huang, Dong, Wei, Lin, Tang, Wang, Yang, Tu, Zhang, Ma, Yang, Xu, Zhou, Bai, He, Lin, Dang, Lu, Chen, Yang, Li, Xue, Ni, Zhang, Wang, Peng, Men, Gao, Lin, Wang, Bai, Tan, Zhu, Li, Liu, Ge, Deng, Zhou, Ren, Zhang, Wei, Ren, Liu, Fan, Yao, Zhang, Wan, Chu, Liu, Cui, Zhang, Guo, and Fan}]{yang2024qwen2technicalreport}
An~Yang, Baosong Yang, Binyuan Hui, Bo~Zheng, Bowen Yu, Chang Zhou, Chengpeng Li, Chengyuan Li, Dayiheng Liu, Fei Huang, Guanting Dong, Haoran Wei, Huan Lin, Jialong Tang, Jialin Wang, Jian Yang, Jianhong Tu, Jianwei Zhang, Jianxin Ma, Jianxin Yang, Jin Xu, Jingren Zhou, Jinze Bai, Jinzheng He, Junyang Lin, Kai Dang, Keming Lu, Keqin Chen, Kexin Yang, Mei Li, Mingfeng Xue, Na~Ni, Pei Zhang, Peng Wang, Ru~Peng, Rui Men, Ruize Gao, Runji Lin, Shijie Wang, Shuai Bai, Sinan Tan, Tianhang Zhu, Tianhao Li, Tianyu Liu, Wenbin Ge, Xiaodong Deng, Xiaohuan Zhou, Xingzhang Ren, Xinyu Zhang, Xipin Wei, Xuancheng Ren, Xuejing Liu, Yang Fan, Yang Yao, Yichang Zhang, Yu~Wan, Yunfei Chu, Yuqiong Liu, Zeyu Cui, Zhenru Zhang, Zhifang Guo, and Zhihao Fan. 2024.
\newblock \href {https://arxiv.org/abs/2407.10671} {Qwen2 technical report}.
\newblock \emph{Preprint}, arXiv:2407.10671.

\bibitem[{Yu et~al.(2024)Yu, Jiang, Shi, YU, Liu, Zhang, Kwok, Li, Weller, and Liu}]{yu2024metamath}
Longhui Yu, Weisen Jiang, Han Shi, Jincheng YU, Zhengying Liu, Yu~Zhang, James Kwok, Zhenguo Li, Adrian Weller, and Weiyang Liu. 2024.
\newblock \href {https://openreview.net/forum?id=N8N0hgNDRt} {Metamath: Bootstrap your own mathematical questions for large language models}.
\newblock In \emph{The Twelfth International Conference on Learning Representations}.

\bibitem[{Yue et~al.(2023)Yue, Chen, Wang, Li, Shen, Liu, Zhou, Xiao, Yun, Huang, and Wei}]{yue2023disclawllm}
Shengbin Yue, Wei Chen, Siyuan Wang, Bingxuan Li, Chenchen Shen, Shujun Liu, Yuxuan Zhou, Yao Xiao, Song Yun, Xuanjing Huang, and Zhongyu Wei. 2023.
\newblock \href {https://arxiv.org/abs/2309.11325} {Disc-lawllm: Fine-tuning large language models for intelligent legal services}.
\newblock \emph{Preprint}, arXiv:2309.11325.

\bibitem[{Zheng et~al.(2024)Zheng, Wang, Ji, Huang, and Peng}]{zheng2024weaktostrong}
Chujie Zheng, Ziqi Wang, Heng Ji, Minlie Huang, and Nanyun Peng. 2024.
\newblock \href {https://openreview.net/forum?id=DbyqbL4ztr} {Weak-to-strong extrapolation expedites alignment}.
\newblock In \emph{ICML 2024 Workshop on Models of Human Feedback for AI Alignment}.

\bibitem[{Zhong et~al.(2019)Zhong, Xiao, Tu, Zhang, Liu, and Sun}]{zhong2019jecqa}
Haoxi Zhong, Chaojun Xiao, Cunchao Tu, Tianyang Zhang, Zhiyuan Liu, and Maosong Sun. 2019.
\newblock \href {https://arxiv.org/abs/1911.12011} {Jec-qa: A legal-domain question answering dataset}.
\newblock \emph{Preprint}, arXiv:1911.12011.

\end{thebibliography}

\appendix

\section{Hyper-Parameter Selection}
\label{sec:appendixa}

For model extrapolation, we empirically choose the hyper-parameter $\mu$ from 1e-4 to 0.8 by logarithmic interval, within $\{1,\,2,\,4,\,6,\,8\}\times\{10^{-4},\,10^{-3},\,10^{-2},\,10^{-1}\}$. 
And for contrastive decoding, we choose the hyper-parameter $\lambda$ within $\{0.1,\, 0.2,\, 0.4,\, 0.6,\,0.8,\,1.0\}$.

For all experiments, we search the optimal hyper-parameters on development sets, and then employ the same hyper-parameters to evaluate models on hold-out test sets.

\begin{table}[t]
\small
\tabcolsep=0.30em
\centering
\begin{tabular}{lcccc}
\toprule
\textbf{Magnitude of $\mu$} & \textbf{DS-7B} & \textbf{QW-1.5B} & \textbf{QW-7B} & \textbf{LM-3B} \\
\midrule
0 (vanilla finetuned) & 57.22 & 52.68 & 66.67 & 53.45\\
1e-4$\sim$8e-4 & \underline{57.58} & 52.87 & \textbf{67.19} & 54.00\\
1e-3$\sim$8e-3 & \textbf{58.39} & \textbf{53.43} & \underline{66.90} & \textbf{55.11}\\
1e-2$\sim$8e-2 & 57.29 & \underline{52.99} & 66.82 & \underline{55.10}\\
1e-1$\sim$8e-1 & 57.06 & 51.38 & 66.53 & 54.20\\
\bottomrule
\end{tabular}
\caption{Performance of extrapolated LLMs on Logic under different $\mu$. The highest results are made \textbf{bold}, with the second \underline{underlined}. DS-7B, QW-1.5B, QW-7B, and LM-3B share the same mearnings with those in Table~\ref{tab:significance_test}.}
\label{tab:mu_effect}
\end{table}

\begin{table}[t]
\small
\tabcolsep=0.30em
\centering
\begin{tabular}{lcccc}
\toprule
\textbf{Magnitude of $\lambda$} & \textbf{DS-7B} & \textbf{QW-1.5B} & \textbf{QW-7B} & \textbf{LM-3B} \\
\midrule
0 (only using ME) & \underline{58.89} & 53.63 & 67.60 & 55.11\\
0.1 & 58.86 & 53.64 & 67.59 & 56.27\\
0.2 & \textbf{58.97} & 53.57 & \textbf{68.06} & 56.68\\
0.4 & 58.77 & \textbf{53.79} & \textbf{68.03} & 56.80\\
0.6 & 58.46 & \underline{53.74} & 67.97 & \underline{57.16}\\
0.8 & 58.55 & 53.48 & 67.81 & \textbf{57.33}\\
1.0 & 58.34 & 52.98 & 67.49 & 57.06\\
\bottomrule
\end{tabular}
\caption{Performance of contrastive decoding on Logic under different $\lambda$. The highest results are made \textbf{bold}, with the second \underline{underlined}. DS-7B, QW-1.5B, QW-7B, and LM-3B share the same mearnings with those in Table~\ref{tab:significance_test}.}
\label{tab:lambda_effect}
\end{table}

Since EpiCoDe can be regarded as a two-stage method, we do NOT use grid search to concurrently optimize $\mu$ and $\lambda$.
Instead, we first select the optimal $\mu$ in model extrapolation. 
We then search $\lambda$ with $\mu$ frozen, ensuring all experiments are fair between EpiCoDe and only using model extrapolation.

We find that these hyper-parameters have similar effects across distinct LLMs and runs. As $\mu$ or $\lambda$ increases from small to large, the improvement brought by model extrapolation or contrastive decoding exhibits convexity. 
For example, Table~\ref{tab:mu_effect} shows the effect of $\mu$ with different magnitudes in model extrapolation on the Logic task~(average over 10 runs). 
We also use Figure~\ref{tab:lambda_effect} to illustrate the effect of different $\lambda$ in contrastive decoding on Logical Reasoning (average over 10 runs).
Similar to previous works~\cite{obrien2023contrastive,zheng2024weaktostrong}, when $\mu$ or $\lambda$ is small, both model extrapolation and contrastive decoding can bring more improvement as the $\mu$ or $\lambda$ grows larger. But if we use very large $\mu$ or $\lambda$, the two methods will fail to improve model performance.

\section{LLMs with Different Sizes in CD}

Inspired by previous work~\cite{li-etal-2023-contrastive,obrien2023contrastive} which employs a small and a large model in contrastive decoding, we also wonder during inference, whether EpiCoDe is still effective if using two models with different numbers of parameters.
Thus, we employ the extrapolated Qwen2-7B-Instruct as the strong model and the finetuned Qwen2-1.5B-Instruct as the weak model.

Table~\ref{tab:different_weak} shows that contrastive decoding can sometimes be effective when we employ a smaller LLM, which is finetuned on the same data, as the weak model. 
However, such setting brings less improvement than EpiCoDe where we employ the same-size finetuned LLM as weak model.

Using LLMs of different sizes is also not robust enough to enhance model performance. For example, in two runs on Math, if we employ  $\theta^{ep}$ of Qwen2-1.5B as the weak model, the accuracy scores decrease from around 57\% to around 16\%, resulting in catastrophic deterioration.

From these additional results, we find that the effectiveness of contrastive decoding may be highly dependent on the \textbf{locality}. Beyond existing methods which employ a large LLM as the strong model and a small LLM as the weak model, our proposed EpiCoDe can achieve the optimal performance.

\begin{table}[t]
\small
\tabcolsep=0.30em
\centering
\begin{tabular}{lccc}
\toprule
\textbf{Weak Model} & \textbf{Law} & \textbf{Math} & \textbf{Logic} \\
\midrule
$\theta^{ft}$ of Qwen2-1.5B & 69.60$_{\text{(+0.57)}}$ & 49.39$_{\text{(\,-\,7.73)}}$ & 67.91$_{\text{(+1.24)}}$\\
\midrule
$\theta^{init}$ of Qwen2-7B & 68.70$_{\text{(\,-\,0.33)}}$ & 55.96$_{\text{(\,-\,1.16)}}$ & 66.78$_{\text{(+0.11)}}$\\
$\theta^{early}$ of Qwen2-7B & 69.63$_{\text{(+0.60)}}$ & 58.27$_{\text{(+1.15)}}$ & 66.75$_{\text{(+0.08)}}$\\
$\theta^{ft}$ of Qwen2-7B & 70.25$_{\text{(+1.22)}}$ & 58.71$_{\text{(+1.59)}}$ & 68.07$_{\text{(+1.40)}}$\\
\bottomrule
\end{tabular}
\caption{The results of different weak models in EpiCoDe. For all experiments, we employ $\theta^{ep}$ of Qwen2-7B-Instruct as the strong model. }
\label{tab:different_weak}
\end{table}

\end{document}